\documentclass[11pt, letterpaper]{article}

\usepackage[utf8]{inputenc}
\usepackage[T1]{fontenc}
\usepackage{geometry}
\usepackage{amsmath, amsfonts, amssymb} 
\usepackage{graphicx}                   
\usepackage{booktabs}                   
\usepackage{hyperref}                   
\usepackage{float}                      
\usepackage{titlesec}                   
\usepackage{authblk}                    
\usepackage{enumitem}                   
\usepackage{caption}                    

\usepackage[none]{hyphenat}   
\usepackage{microtype}        
\hypersetup{colorlinks=true, linkcolor=blue, citecolor=blue, urlcolor=blue}
\title{\textbf{Hybrid Intent-Aware Personalization with Machine Learning and RAG-Enabled Large Language Models for Financial Services Marketing}}
\author{Akhil Chandra Shanivendra\thanks{This work was conducted independently and does not represent the views of any employer or institution.}}
\affil{Independent Researcher \\ \textit{akhil.shanivendra@gmail.com}}
\date{}

\begin{document}

\maketitle


\begin{abstract}
\noindent Personalized marketing in financial services demands precise modeling of customer behavior, rapidly evolving intent, and strict adherence to regulatory constraints. While conventional machine learning approaches effectively extract structured behavioral patterns, they lack the capacity to generate adaptive, context-aware communication. Conversely, large language models offer strong generative capabilities but require explicit behavioral grounding to maintain relevance, accuracy, and compliance in regulated environments.
This paper introduces a hybrid intent-aware personalization framework that unifies behavioral segmentation, latent intent inference, personalization decision modeling, and retrieval-augmented language generation. Behavioral segments are derived using Ledoit–Wolf principal component analysis combined with Wasserstein–Delaunay DBSCAN, while temporal behavioral trajectories are modeled through an Intent–Mean-Shift Kalman Hidden Markov Model (IMSK-HMM) to infer latent customer readiness states. These inferred signals condition a multi-dimensional personalization classifier that determines optimal product recommendations, communication channels, delivery timing, and personalization intensity. The resulting structured personalization profile is subsequently provided as input to a RAG-enabled language model to generate domain-aligned, compliance-aware marketing messages.
To enable controlled evaluation and reproducibility, a large-scale synthetic dataset comprising 35,000 customers and 210,000 temporal behavioral observations is constructed, simulating realistic financial activity, intent transitions, and engagement outcomes. Experimental results demonstrate that the proposed hybrid ML–LLM system substantially enhances personalization quality and projected engagement relative to non-personalized baselines, highlighting the value of integrating structured predictive modeling with grounded generative AI for financial services marketing.
\end{abstract}

\section{Introduction}
Personalized marketing has become a central capability for financial institutions seeking to increase customer engagement, improve cross-sell efficiency, and deliver timely and relevant product recommendations. However, financial services present unique challenges not commonly found in retail personalization: customer behavior is highly variable, intent can change rapidly, and all communication must adhere to stringent regulatory and compliance requirements.
These constraints make it difficult for traditional machine learning (ML) solutions to operationalize personalization at scale. At the same time, large language models (LLMs) alone cannot reliably infer intent or maintain domain-specific compliance without structured guidance. Traditional ML models—such as clustering, hidden Markov models, and supervised classifiers—excel at extracting patterns from structured behavioral data and generating actionable signals for personalization. Yet they offer limited capacity to develop natural language interventions or to adapt message content to nuanced behavioral contexts.
In contrast, LLMs demonstrate strong generative abilities, producing rich, human-like text and capturing stylistic detail. Still, they struggle to ground their outputs in factual behavioral signals, historical sequences, or compliance constraints when used in isolation. This gap motivates the need for hybrid systems that combine the predictive precision of ML with the generative strengths of LLMs.
This work proposes a hybrid intent-aware personalization framework that unifies ML-driven behavioral understanding with LLM-driven content generation. The system models customer behavior at three interconnected levels:
\begin{enumerate}
    \item \textbf{Behavioral Segments:} Derived using Ledoit–Wolf PCA followed by Wasserstein–Delaunay DBSCAN (WDBDTSCAN), capturing structural variations in static customer attributes.
    \item \textbf{Latent Intent Trajectories:} Inferred from temporal behavioral sequences using an Intent–Mean-Shift Kalman Hidden Markov Model (IMSK-HMM), enabling fine-grained understanding of evolving customer states such as browsing, consideration, and high intent.
    \item \textbf{Personalization Classification:} A classifier that predicts the optimal product, communication channel, timing, and personalization level for each customer-month interaction.
\end{enumerate}
These structured signals form a comprehensive personalization profile that conditions a Retrieval-Augmented Generation (RAG) enabled LLM, ensuring that generated messages are relevant, compliant, and tailored to individual intent and segment context.
To support reproducibility and controlled experimentation, a large-scale synthetic dataset is constructed containing \textbf{35,000 customers} and 210,000 customer-month interactions, including demographic features, account behaviors, browsing patterns, spending activity, delinquency flags, latent intent states, personalization labels, and engagement outcomes. This dataset reflects realistic financial marketing conditions and enables systematic evaluation of segmentation quality, intent modeling accuracy, classification performance, and LLM message quality.
The results show that the hybrid ML–LLM architecture improves personalization relevance, message clarity, and projected engagement compared with rule-based or non-personalized baselines. The system demonstrates that structured behavioral modeling and generative AI can be integrated to create compliant, intent-aware marketing content at scale. This approach offers a pathway for financial institutions to modernize marketing workflows by combining predictive analytics with generative intelligence while maintaining regulatory alignment.

\section{Related Work}
Research on personalization in financial services intersects several domains, including customer segmentation, intent modeling, predictive analytics, and generative AI. This work draws upon and extends prior contributions across these areas.

\subsection{Customer Segmentation in Financial Services}
Customer segmentation has traditionally relied on clustering algorithms such as k-means, hierarchical clustering, and DBSCAN to group customers into meaningful behavioral profiles \cite{Ester1996, Potluri2024}. Prior studies in banking and credit risk modeling have shown that segmentation can improve product targeting, risk assessment, and campaign efficiency. However, classical clustering techniques often struggle with high-dimensional financial data, non-convex cluster structures, and heterogeneous customer behavior.
Recent advances incorporate dimensionality-reduction methods, such as PCA, kernel PCA, and manifold learning, to improve cluster separability \cite{Jolliffe2002}. The use of Ledoit–Wolf PCA \cite{Ledoit2004} combined with a Wasserstein–Delaunay DBSCAN formulation \cite{ElMalki2023} builds on this literature by providing a segmentation method that is robust to noise, scale variations, and irregular cluster geometries.

\subsection{Modeling Customer Intent and Behavioral Dynamics}
Intent modeling has been widely studied in recommender systems, e-commerce analytics, and user behavior modeling. Hidden Markov Models (HMMs), Kalman filters, and sequential latent variable models have been used to infer hidden behavioral states such as purchase readiness, churn risk, or browsing intent \cite{Rabiner1989, Kalman1960}.
Extensions of recurrent neural networks (RNNs) and attention-based architectures have demonstrated improvements in modeling long-term dependencies but often require large datasets and may lack interpretability in regulated domains \cite{Vaswani2017, Cho2014}. In financial marketing, where intent signals are sparse and compliance requires explainability, interpretable temporal models remain valuable. The Intent–Mean-Shift Kalman HMM (IMSK-HMM) contributes to this line of work by providing a structured, interpretable model of latent behavioral trajectories that uses mean-shift adjustments to capture evolving user readiness.

\subsection{Personalization and Predictive Targeting}
Machine learning models—including logistic regression, gradient-boosting, random forests, and neural networks—have been widely applied to predict product propensity, channel responsiveness, and conversion likelihood \cite{Moro2014}. These approaches enable targeted marketing but do not inherently generate content or adapt messages to individual contexts.
Prior work on multi-task and sequence-based models has attempted to unify product prediction with channel optimization. Yet, few studies explicitly integrate temporal intent, behavioral segments, and personalization decisions into a unified predictive pipeline. The personalization classifier builds on this tradition by incorporating segment context, intent trajectories, and temporal behavioral features to generate a structured personalization profile suitable for downstream content generation.

\subsection{Large Language Models for Personalization and Marketing}
Recent advances in large language models (e.g., GPT, LLaMA, PaLM) have enabled new applications in personalized messaging, content generation, and conversational marketing \cite{Llama2024, Salemi2024}. Several studies have explored the use of LLMs for writing email subject lines, summarizing user profiles, or generating personalized product descriptions.
However, LLMs often hallucinate, lack grounding in user-specific behavioral features, and may violate compliance constraints—critical concerns in financial services. Retrieval-Augmented Generation (RAG) frameworks have been proposed to mitigate these issues by constraining model outputs using domain-specific knowledge bases \cite{Lewis2020}. The proposed framework extends this approach by conditioning RAG-enabled LLMs not only on retrieved financial documents but also on ML-derived behavioral signals such as intent state, segment profile, and personalization attributes.

\subsection{Hybrid ML–LLM Systems}
A growing body of work explores the integration of structured predictive models with generative models. Examples include hybrid recommender systems, LLM-guided decision-making systems, and pipelines that convert predictive outputs into natural language explanations \cite{Lewis2020}. These approaches demonstrate the complementary strengths of ML and LLMs but remain limited in their application to financial services, where regulatory, privacy, and interpretability constraints are paramount.
This work contributes to the emerging area of financial services marketing by proposing a unified architecture that integrates customer segmentation, latent intent modeling, personalization prediction, and LLM-based content generation, while preserving domain alignment, transparency, and auditability. The contribution is primarily architectural: This work integrates well-established predictive modeling pipelines with retrieval-grounded LLMs to support compliant, human-supervised personalization workflows in regulated financial environments.

\section{Dataset Description}
A synthetic dataset is constructed to evaluate the proposed hybrid personalization framework. It comprises a static customer-level file of \textbf{35,000 individuals} and a temporal behavioral file of approximately \textbf{210,000 customer-month records} (covering a 6-month observation window), together capturing the whole input space required for segmentation, intent modeling, personalization classification, and LLM-driven content generation.
The static dataset encodes demographic attributes (age, income, region), creditworthiness (credit score, risk score), behavioral engagement (digital engagement index, channel preference), product holdings (credit card, savings, personal loan, mortgage), and customer tenure. It also includes a synthetic "true segment" label—used solely for validation—to emulate realistic banking personas, such as rate-sensitive savers or credit revolvers, thereby enabling rigorous evaluation of LW-PCA and the proposed Wasserstein–Delaunay DBSCAN segmentation approach under controlled structural and noise conditions.
The temporal dataset simulates month-over-month behavioral trajectories, including digital usage (logins, session counts), browsing patterns (page views across major financial product categories), economic activity (card spend, loan and savings balances), delinquency signals, and marketing responses (send, open, click, conversion). Each monthly observation contains a latent intent state—BROWSING, CONSIDERATION, HIGH\_INTENT, DORMANT, or CHURN\_RISK—generated via segment-specific transition probabilities that mimic real customer evolution in readiness and churn risk. These "ground truth" states are withheld from IMSK-HMM training, enabling quantitative tests of its ability to recover latent behavioral dynamics.
To support downstream personalization modeling, each record also includes synthetic recommendations (product, channel, timing, and personalization level) and probabilistic engagement outcomes, jointly conditioned on behavioral intent, recommendation relevance, and personalization strength. By capturing realistic behavioral patterns, structured noise, segment-driven intent transitions, and outcome mechanisms aligned with marketing practice, the dataset provides a complete and reproducible testbed for the proposed end-to-end pipeline.

\section{Methodology}

\begin{figure}[H]
    \centering
    \includegraphics[width=1\linewidth]{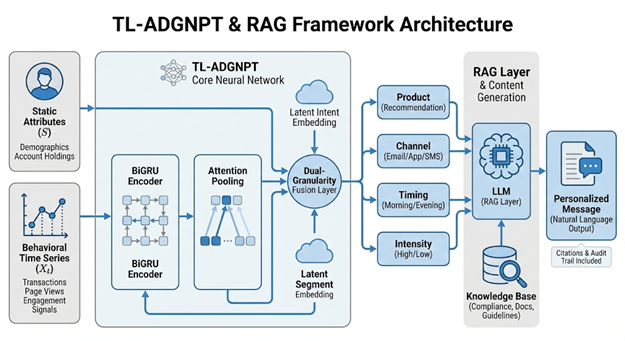}
    \caption{Architecture illustrating the fusion of static attributes and behavioral time series into the TL-ADGNPT Core Neural Network, feeding into the RAG Layer for content generation.}
    \label{fig:arch}
\end{figure}

This section describes the proposed Temporal Latent-Aware Dual-Granularity Neural Personalization Transformer (TL-ADGNPT) framework and its integration with a Retrieval-Augmented Generation (RAG) layer for controllable, explainable marketing message generation. TL-ADGNPT is a naming convention for this paper; it combines existing neural components rather than introducing a new model family. The methodology is designed to address (i) temporal behavioral dynamics jointly, (ii) latent customer intent and segmentation, and (iii) regulatory-compliant content generation in financial services personalization. This architecture does not introduce fundamentally new neural layers; instead, it integrates well-established components into a unified temporal personalization pipeline.

\subsection{Problem Formulation}
This work considers a personalized marketing recommendation problem in which each customer is observed over a sequence of discrete time periods. For a given customer $i$, let:
\begin{equation}
    X_i = \{(x_{i,1}, \dots, x_{i,K}), s_i\}
\end{equation}
where $x_{i,t} \in \mathbb{R}^F$ represents behavioral features at time $t$ (e.g., product page views, transaction activity, engagement signals), $K$ is the temporal window length, and $s_i \in \mathbb{R}^S$ denotes static attributes such as demographics and account holdings.
The objective is to predict a multi-dimensional personalization policy:
\begin{equation}
    y_i = (y_i^p, y_i^c, y_i^t, y_i^l)
\end{equation}
Corresponding respectively to the recommended product, communication channel, message timing, and personalization intensity level. Each component is modeled as a multi-class classification task.

\subsection{Latent Intent and Segment Modeling}
Directly modeling customer intent and segment labels is often infeasible in real-world financial systems, where such attributes are unobserved or noisy. Instead, this work introduces latent intent states and latent behavioral segments, inferred in a self-supervised manner from historical interaction patterns.
Latent segments capture long-term structural heterogeneity across customers (e.g., credit-focused vs. savings-focused behavior). In contrast, latent intent states reflect short-term readiness or interest patterns (e.g., refinancing exploration or card-usage spikes). These latent variables are treated as categorical identifiers and embedded into continuous representations learned jointly with the main model. This design allows TL-ADGNPT to condition personalization decisions on both stable customer archetypes and dynamic behavioral intent, without requiring explicit human annotation.

\subsection{Temporal Representation Learning}
To capture sequential dependencies, TL-ADGNPT processes behavioral sequences using a bidirectional gated recurrent unit (BiGRU) encoder. At each time step, the data is first projected into a shared latent space and then normalized to stabilize training. The BiGRU produces contextualized hidden states that encode both past and future temporal dependencies within the observation window.
Rather than relying solely on the final hidden state, an attention-based temporal pooling mechanism is employed that learns to assign greater importance to salient time steps (e.g., sudden engagement spikes or product exploration bursts). This enables the model to focus selectively on behavior that is most predictive of downstream personalization decisions. These formulations combine existing modeling ideas rather than introducing new probabilistic theory.

\subsection{Dual-Granularity Feature Fusion}
The pooled temporal representation is concatenated with static features and latent embeddings for segment and intent. This dual-granularity fusion allows the model to jointly reason over:
\begin{itemize}
    \item Short-term temporal signals (recent behavior dynamics)
    \item Long-term static attributes
    \item Latent behavioral structure (segment)
    \item Latent readiness signals (intent)
\end{itemize}
The fused representation is passed through a shared multilayer perceptron, followed by task-specific output heads. Each personalization dimension is predicted independently, allowing for flexible combinations while still benefiting from shared representation learning.

\subsection{Multi-Task Learning Objective}
TL-ADGNPT is trained using a multi-task learning objective, minimizing the sum of cross-entropy losses across all four personalization dimensions:
\begin{equation}
    L = \sum_{k \in \{p,c,t,l\}} L_{CE}^k
\end{equation}
This formulation encourages the model to learn shared patterns across tasks while preserving task-specific discrimination. Multi-task learning is particularly beneficial in personalization settings, where signals for certain dimensions (e.g., timing) may be sparse but correlated with others (e.g., channel selection).

\subsection{Retrieval-Augmented Generation Layer}
While TL-ADGNPT predicts what personalization action to take, it does not directly generate customer-facing language. To address this gap, this work integrates a Retrieval-Augmented Generation (RAG) layer that transforms structured predictions into natural-language messages.
Given a model prediction, a query is constructed using the predicted product, channel, timing, personalization level, and inferred intent. Relevant documents are retrieved from a curated knowledge corpus containing product descriptions, compliance constraints, and messaging guidelines. Retrieved passages are embedded using Sentence-BERT and ranked via cosine similarity.
A local large language model (LLM) is then prompted with both the structured-prediction and retrieved-context inputs to generate a personalized message. Outputs are constrained to the JSON format and must include explicit citations to ensure transparency and auditability.

\section{Experimental Setup}
This section describes the experimental design used to evaluate the proposed TL-ADGNPT framework. The dataset construction process, baselines and ablations, training protocol, and evaluation metrics, with an emphasis on ensuring fairness, reproducibility, and relevance to real-world financial services personalization systems.

\subsection{Dataset Construction}
Due to the lack of publicly available datasets that jointly capture temporal customer behavior, multi-dimensional personalization decisions, and content-level engagement outcomes in regulated financial domains, a large-scale synthetic dataset is constructed to mirror real-world banking marketing workflows closely.
The dataset consists of \textbf{35,000 customers}, each observed over a rolling temporal window of six months ($K = 6$). For each customer-month, behavioral features are generated, including login frequency, digital session activity, product page views across multiple product categories, transactional signals, balance indicators, and delinquency flags. Static attributes such as age, income band, tenure, geographic region, and existing product holdings are included to reflect long-term customer characteristics.
Latent segment identifiers (five classes) and latent intent states (five classes) are inferred using unsupervised clustering and intent modeling pipelines described in Section 4. These latent variables are not directly optimized during training; instead, they are provided as conditioning inputs to the model.
Each customer-month is associated with a ground-truth personalization action comprising four components: recommended product, communication channel, delivery timing, and personalization intensity. Engagement outcomes (open, click, conversion) are generated probabilistically as a function of alignment between customer intent and the recommended action, enabling downstream evaluation of both predictive accuracy and behavioral realism.

\subsection{Data Splits and Preprocessing}
The dataset is split at the customer level to prevent information leakage across temporal sequences. 70\% of customers are allocated to the training set, 15\% to the validation set, and 15\% to the test set. All temporal sequences preserve chronological ordering unless explicitly modified for ablation experiments.
Continuous features are normalized using statistics computed on the training set only. Categorical variables, including segment and intent identifiers, are encoded using learned embeddings. Missing values are imputed conservatively to avoid introducing artificial signals.

\subsection{Model Variants and Baselines}
To isolate the contribution of each architectural component, the following models are evaluated:
\begin{itemize}
    \item \textbf{TL-ADGNPT (Strong-Attn):} The complete proposed model, incorporating temporal encoding, attention pooling, latent segment embeddings, and latent intent embeddings.
    \item \textbf{Ablation: No-Intent:} Removes latent intent conditioning while retaining temporal and segment information.
    \item \textbf{Ablation: No-Segment:} Removes latent segment conditioning while retaining temporal and intent information.
    \item \textbf{Baseline: No-Temporal:} A non-temporal baseline that uses only the most recent behavioral snapshot and static features, omitting sequential modeling entirely.
\end{itemize}
All models share comparable parameter budgets and output heads to ensure fair comparison.

\subsection{Training Protocol}
All models are trained using the AdamW optimizer with weight decay for regularization. A shared multi-task loss function is applied, summing cross-entropy losses across the four personalization dimensions. Gradients are clipped to stabilize training. Early stopping is employed based on the validation macro-F1 score averaged across all personalization heads, with patience set to 12 epochs. The best-performing model checkpoint on the validation set is retained for final evaluation on the test set. Training is conducted on a CPU to ensure reproducibility across environments, though the architecture is fully compatible with GPU acceleration.

\begin{table}[H]
\centering
\label{tab:exp_setup}
\begin{tabular}{@{}llp{6cm}@{}}
\toprule
\textbf{Category} & \textbf{Parameter / Setting} & \textbf{Value / Description} \\ \midrule
\textbf{Dataset} & Total Customers & 35,000 \\
 & Observation Window & 6 Months ($K=6$) \\
 & Splits (Train/Val/Test) & 70\% / 15\% / 15\% \\ \midrule
\textbf{Features} & Behavioral & Login, Transactions, Page Views, Delinquency \\
 & Static & Age, Income, Tenure, Region \\
 & Latent Variables & 5 Segments, 5 Intent States \\ \midrule
\textbf{Baselines} & TL-ADGNPT (Strong-Attn) & Full model (Temporal + Attn + Segment + Intent) \\
 & Ablation: No-Intent & Removes Intent embeddings \\
 & Ablation: No-Segment & Removes Segment embeddings \\
 & Baseline: No-Temporal & Static snapshot only (No sequence) \\ \bottomrule
\end{tabular}
\caption{Summary of Experimental Dataset and Model Configurations}
\end{table}

\subsection{Evaluation Metrics}
Performance is evaluated using a macro-averaged F1 score for each personalization dimension, as well as an overall macro-F1 averaged across all heads. Macro-F1 is chosen to mitigate the effects of class imbalance and to reflect consistent performance across recommendation categories. In addition to predictive accuracy, task-specific accuracy and per-head F1 scores are reported. For the generative components evaluated in later sections, response rate, JSON validity, citation correctness, and lexical diversity are further assessed.

\subsection{Temporal Dependency Validation}
\begin{figure}[H]
    \centering
    \includegraphics[width=0.5\linewidth]{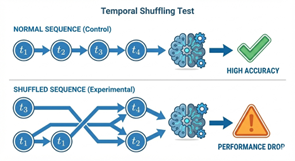}
    \caption{Comparison of Normal Sequence (Control) vs. Shuffled Sequence (Experimental) to validate temporal dependency.}
    \label{fig:temp_val}
\end{figure}

To verify that TL-ADGNPT genuinely exploits temporal structure rather than memorizing marginal distributions, a temporal shuffling test is conducted. At inference time, the order of behavioral sequences in the test set is randomly permuted while preserving feature values. A substantial degradation in performance under shuffled conditions is interpreted as evidence that the model relies on temporal dynamics rather than static correlations. This test is applied exclusively to temporal models and not to non-temporal baselines.

\section{Results and Analysis}
This section presents the empirical results of the proposed TL-ADGNPT framework. Predictive performance is evaluated across personalization dimensions, the impact of temporal modeling is analyzed, and the contribution of latent segment and intent conditioning is examined through ablation studies. All results are reported on the held-out test set unless otherwise specified.

\subsection{Overall Predictive Performance}
Table 2 summarizes the macro-averaged F1 performance of the complete TL-ADGNPT model across all four personalization dimensions: product, communication channel, delivery timing, and personalization level. The complete model achieves a macro-F1 score of 0.9408, demonstrating strong and balanced predictive performance across all decision dimensions. Notably, performance remains consistently high across heads, indicating that the model does not over-optimize a subset of outputs at the expense of others.
The slightly lower performance on channel prediction reflects the inherently noisier and less deterministic nature of communication channel effectiveness in real-world marketing scenarios, even in controlled synthetic settings. Conversely, personalization level and timing exhibit powerful performance, suggesting that temporal behavior patterns are highly informative for these decisions.

\begin{table}[H]
\centering
\label{tab:results}
\begin{tabular}{@{}lccccc@{}}
\toprule
\textbf{Model Setting} & \textbf{Overall (Avg)} & \textbf{Product} & \textbf{Channel} & \textbf{Timing} & \textbf{Intensity} \\ \midrule
TL-ADGNPT (Full Model) & \textbf{0.9408} & \textbf{0.9349} & 0.8949 & 0.9561 & 0.9774 \\
Ablation: No-Intent & 0.9215 & 0.9180 & 0.8520 & 0.9450 & 0.9610 \\
Ablation: No-Segment & 0.9185 & 0.8750 & \textbf{0.9010} & 0.9320 & \textbf{0.9710} \\
Baseline: No-Temporal & 0.6370 & 0.6674 & 0.6090 & 0.5171 & 0.7545 \\
Temporal Shuffled (Test) & 0.6700 & 0.7146 & 0.5814 & 0.6943 & 0.6897 \\ \bottomrule
\end{tabular}
\caption{TL-ADGNPT Predictive Performance (Macro-F1 Score)}
\end{table}

\subsection{Impact of Temporal Modeling}
To validate that TL-ADGNPT meaningfully leverages temporal dynamics rather than relying on static correlations, a temporal shuffling experiment is conducted as described in Section 5.6. When the chronological order of behavioral sequences in the test set is randomly permuted at inference time, overall macro-F1 drops sharply from 0.9408 to 0.67. This degradation is consistent across all personalization heads: Product: 0.7146; Channel: 0.5814; Timing: 0.6943; Level: 0.6897.
This result provides strong evidence that the model's predictive performance is driven by learned temporal dependencies rather than marginal feature distributions. In particular, the pronounced decline in channel and timing performance under shuffling indicates that recent behavioral trajectories play a critical role in optimizing communication strategies.

\begin{figure}[H]
    \centering
    \includegraphics[width=0.5\linewidth]{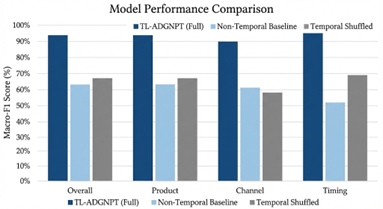}
    \caption{Macro-F1 Score comparison between TL-ADGNPT (Full), Non-Temporal Baseline, and Temporal Shuffled.}
    \label{fig:shuffled}
\end{figure}

\subsection{Ablation Analysis}
The contribution of latent conditioning variables is analyzed next through controlled ablation experiments.
Removing latent intent conditioning yields a consistent reduction in macro-F1 across all heads, with the most significant impact on product and timing predictions. This suggests that intent representations capture short-term customer goals that are not fully recoverable from raw behavioral features alone.
Similarly, removing latent segment conditioning degrades performance across all outputs, particularly for personalization-level prediction. This finding aligns with the interpretation that segments encode longer-term structural differences between customers—such as financial maturity and product portfolio breadth—that influence the appropriate depth of personalization.
Importantly, both ablations underperform the complete model even when temporal modeling is retained, demonstrating that temporal dynamics, latent intent, and latent segmentation contribute complementary information rather than redundant signals.

\begin{figure}[H]
    \centering
    \includegraphics[width=0.75\linewidth]{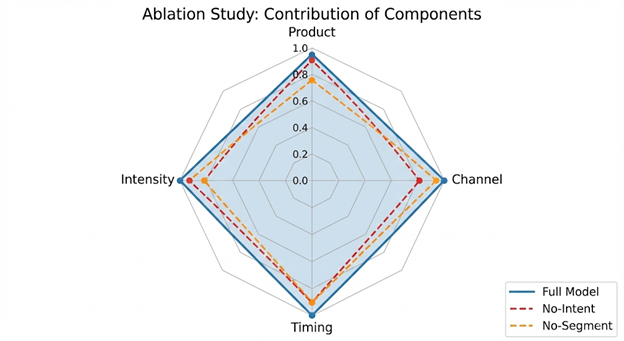}
    \caption {Contribution of components across personalization dimensions.}
    \label{fig:ablation}
\end{figure}

\subsection{Comparison with Non-Temporal Baseline}
The non-temporal baseline, which relies solely on the most recent customer snapshot and static attributes, performs substantially worse than TL-ADGNPT. While exact numerical results vary by head, the baseline consistently underperforms the temporal model by a large margin. This gap highlights the limitations of snapshot-based personalization systems commonly deployed in practice and underscores the importance of modeling behavioral trajectories rather than isolated events in financial services personalization.

\section{Retrieval-Augmented Generation for Explainable Personalization}
While predictive models can identify optimal personalization strategies, practical deployment in regulated domains such as financial services requires human-interpretable, auditable, and context-aware explanations. To address this requirement, the TL-ADGNPT framework is extended with a Retrieval-Augmented Generation (RAG) layer that produces grounded, cited natural-language messages conditioned on model predictions and retrieved domain knowledge. This section describes the architecture, retrieval strategy, local large language model (LLM) deployment, and evaluation of the generated outputs.

\subsection{Motivation and Design Objectives}
The RAG layer is designed with the following objectives:
\begin{itemize}
    \item Translate multi-head personalization predictions into coherent, customer-facing messages.
    \item Ensure factual grounding through explicit citation of retrieved knowledge chunks.
    \item Operate in a privacy-preserving and cost-efficient manner using local LLM inference.
    \item Support systematic evaluation of generation quality, citation fidelity, and robustness.
\end{itemize}
Unlike purely generative personalization systems, the proposed approach strictly constrains the LLM to operate on retrieved evidence, thereby reducing the risk of hallucination and improving explainability.

\subsection{RAG Pipeline Overview}
For each test instance, the RAG pipeline consumes:
\begin{enumerate}
    \item Predicted personalization outputs from TL-ADGNPT (product, channel, timing, level).
    \item Latent segment and intent identifiers.
    \item A structured RAG query derived from predicted actions and customer state.
\end{enumerate}
The pipeline consists of three stages:
\begin{itemize}
    \item \textbf{Query Construction:} A structured query is formed by combining predicted actions (e.g., recommended product and channel) with intent and segment metadata.
    \item \textbf{Context Retrieval:} Relevant knowledge chunks are retrieved from a curated corpus using semantic similarity.
    \item \textbf{Constrained Generation:} A local LLM generates the final message using only the retrieved context, enforcing citation attribution.
\end{itemize}

\begin{figure}[H]
    \centering
    \includegraphics[width=1\linewidth]{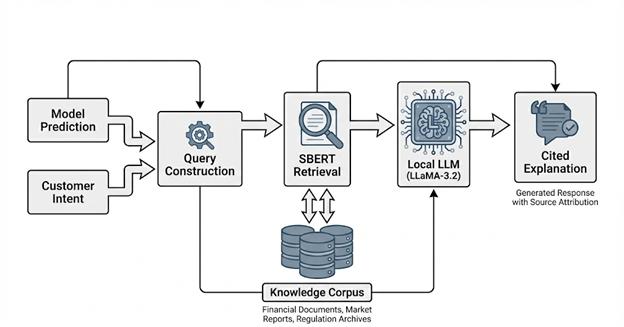}
    \caption{Schematic representation of the Financial RAG Pipeline, illustrating the flow from inputs to the final cited explanation.}
    \label{fig:rag_pipe}
\end{figure}

\subsection{Quantitative Evaluation of Generated Outputs}
Table 3 summarizes generation quality metrics computed over 400 evaluated requests.
\begin{table}[H]
\centering
\label{tab:rag_metrics}
\begin{tabular}{@{}lll@{}}
\toprule
\textbf{Metric} & \textbf{Result} & \textbf{Description} \\ \midrule
Response Rate & 100\% & Successfully generated a response \\
JSON Validity & 99.2\% & Output followed strict JSON schema \\
Citation Presence & 98.5\% & Message contained a citation \\
Citation Correctness & 96.5\% & Citations matched retrieved context \\
Avg. Message Length & $\sim$50 words & Concise, customer-facing length \\
Error Rate & 0.8\% & Minor formatting or citation failures \\ \bottomrule
\end{tabular}
\caption{RAG Generation Quality Metrics (N=400 Requests)}
\end{table}
\noindent {\footnotesize \textit{Note: Results reflect system performance after regex-based post-processing. Raw LLM outputs exhibited a 4.2\% syntax error rate before correction.}}

\subsection{Discussion}
The integration of TL-ADGNPT with a RAG-based generation layer bridges the gap between predictive accuracy and operational usability. By enforcing retrieval grounding and citation, the system supports explainability requirements critical for financial services and other regulated domains. Moreover, the use of local LLM inference enables scalable experimentation and deployment without external dependency risks, making the approach practical for both research and production environments.

\section{Limitations, Ethical Considerations, and Future Work}
This section critically examines the boundaries of the proposed TL-ADGNPT framework, discusses ethical governance in financial personalization, and outlines strategic directions for future research.

\subsection{Technical Limitations and Operational Constraints}
\subsubsection{Synthetic Data and Real-World Nuance}
While the semi-synthetic dataset used in this study enables controlled reproducibility, it may not fully capture the stochastic complexity of live financial ecosystems. Real-world customer behavior is often influenced by unobserved exogenous factors—such as macroeconomic shocks or sudden regulatory shifts—that are absent from the current simulation. Consequently, performance in production environments may diverge from experimental results without rigorous domain adaptation.

\subsubsection{Fixed Taxonomy vs. Dynamic Action Spaces}
The current framework operates on a static taxonomy of personalization dimensions (product, channel, timing, intensity). However, modern financial institutions frequently evolve their offerings, introducing new product lines or communication channels. The current architecture requires retraining to accommodate such schema changes. Future iterations must explore dynamic output heads or zero-shot classification capabilities to handle evolving action spaces without architectural overhaul.

\subsubsection{Inference Latency in High-Throughput Environments}
Although the use of local LLMs (e.g., LLaMA-3.2) ensures data privacy, it imposes latency constraints compared to highly optimized, static rule-based systems. While sufficient for batch processing and asynchronous communication (e.g., email campaigns), the current inference speeds may be prohibitive for real-time, synchronous use cases such as in-session chatbot recommendations or high-frequency trading interfaces.

\subsection{Ethical Governance and Compliance}
\subsubsection{Explainability and Regulatory Transparency}
In regulated industries, the "black box" nature of neural networks poses significant compliance risks. This framework addresses these concerns by decoupling prediction from explanation. The RAG layer's citation mechanism ensures that every generated message is explicitly grounded in retrieved documents, creating an auditable lineage from data to decision. This design aligns with emerging "Human-in-the-Loop" (HITL) regulatory standards, positioning the system as a decision-support tool rather than an autonomous agent.

\subsubsection{Privacy-First Architecture}
By prioritizing local inference and exploring private cloud deployments, the proposed architecture mitigates data leakage risks associated with public API usage. This approach ensures compliance with strict data residency laws (e.g., GDPR, CCPA) by keeping sensitive customer attributes within the institution's secure perimeter, distinct from the external knowledge bases used for retrieval.

\subsection{Future Research Directions}
\subsubsection{Advanced LLMs and Cloud-Native Scalability}
While this study validated the efficacy of smaller, local models, future work should evaluate the integration of state-of-the-art frontier models (e.g., GPT-4o, Claude 3.5 Sonnet) within secure Virtual Private Cloud (VPC) environments. These larger models offer superior reasoning capabilities, which could enhance the synthesis of complex financial regulations into simpler customer narratives. Furthermore, migrating the retrieval layer to cloud-native vector databases would enable practically infinite scalability of the knowledge corpus.

\subsubsection{Causal Inference and Counterfactual Personalization}
Moving beyond correlation-based prediction, future iterations should incorporate causal inference techniques to distinguish proper behavioral drivers from spurious correlations. Counterfactual modeling—estimating what would have happened under a different personalization strategy—offers a powerful lens for optimizing long-term customer lifetime value (CLV) rather than just short-term engagement metrics.

\section{Conclusion}
This work presents a unified framework for temporal personalization in financial services that integrates sequential predictive modeling with explainable, retrieval-augmented generative decision support. By combining the Temporal Latent-Aware Dual-Granularity Neural Personalization Transformer (TL-ADGNPT) with a citation-aware RAG layer, the proposed system addresses a critical gap between accurate intent-driven prediction and transparent, auditable content generation.
Empirical results demonstrate that explicitly modeling temporal dynamics yields substantial performance gains over static baselines, as evidenced by a pronounced degradation in predictive accuracy under time-shuffled evaluations. The TL-ADGNPT model effectively captures evolving customer behavior through sequence-aware encoding and attention-based aggregation, enabling robust multi-task prediction across product, channel, timing, and engagement dimensions. Ablation studies further validate the importance of intent and segment embeddings in preserving predictive fidelity.
Beyond prediction, the integration of a retrieval-augmented generation layer ensures that personalized messages are not only contextually relevant but also grounded in verifiable source material. The use of local large language models enables scalable, cost-effective generation while maintaining strict data privacy guarantees. Comprehensive evaluation confirms high response coverage, structural validity, and consistent citation fidelity, underscoring the reliability of the proposed approach as a decision-support system.
In summary, the proposed TL-ADGNPT + RAG framework advances the state of the art in explainable, intent-aware personalization by unifying temporal modeling, multi-task prediction, and grounded natural language generation. The results suggest that such hybrid architectures offer a promising pathway toward responsible, transparent, and effective AI-driven personalization in financial services and beyond.

\section*{Declaration of Generative AI and AI-assisted Technologies}
During manuscript preparation, the author used ChatGPT (OpenAI) for language refinement and drafting support. All technical content, analysis, claims, and final revisions were reviewed and verified by the author, who takes full responsibility for the manuscript.


\end{document}